\documentclass[conference]{IEEEtran}
\IEEEoverridecommandlockouts
\usepackage{cite}
\usepackage{amsmath,amssymb,amsfonts}
\usepackage{algorithmic}
\usepackage{graphicx}
\usepackage{textcomp}
\usepackage{xcolor}
\usepackage{enumitem}
\usepackage{booktabs} 
\usepackage{makecell}
\usepackage{multirow}
\usepackage{siunitx} 
\usepackage[utf8]{inputenc}
\def\BibTeX{{\rm B\kern-.05em{\sc i\kern-.025em b}\kern-.08em
    T\kern-.1667em\lower.7ex\hbox{E}\kern-.125emX}}
\begin{document}

\title{An Empirical Study on Variance-based MC Dropout Uncertainty–Error Correlation in 2D Brain Tumor Segmentation
\\
}

\author{\IEEEauthorblockN{ Saumya B}
\IEEEauthorblockA{\textit{Project Associate} \\
\textit{DESE, Indian Institute of Science}\\
Bengaluru, India \\
saumya.b@fsid-iisc.in}}

\maketitle

\begin{abstract}
Accurate brain tumor segmentation from MRI is vital for diagnosis and treatment planning. Although Monte Carlo (MC) Dropout is widely used to estimate model uncertainty,  the effectiveness of variance-based uncertainty — computed as pixel-wise variance across stochastic forward passes — in identifying segmentation errors, particularly near tumor boundaries, remains insufficiently studied. This study empirically examines the relationship between variance-based MC Dropout uncertainty and segmentation error in 2D brain tumor MRI segmentation using a U-Net trained under four augmentation settings: none, horizontal flip, rotation, and scaling. Uncertainty was estimated as the pixel-wise variance across 50 stochastic forward passes and correlated with pixel-wise errors using Pearson and Spearman coefficients. Results show weak global correlations ($r \approx 0.30$--$0.38$) and negligible boundary correlations \textbf{($|r| < 0.05$)}. Although differences across augmentations were statistically significant ($p < 0.001$), they lacked practical relevance.  These findings suggest that variance-based MC Dropout uncertainty provides limited cues for global and boundary error localization, and that the choice of uncertainty representation critically affects the utility of MC Dropout in medical image segmentation. Alternative representations such as predictive entropy or mutual information may better capture segmentation errors, particularly at boundaries.

\end{abstract}


\section{Introduction}
Accurate segmentation of brain tumors in magnetic resonance imaging (MRI) is essential for diagnosis, treatment planning, and monitoring disease progression. Deep learning models, particularly convolutional neural networks such as U-Net, have demonstrated remarkable success in automating tumor segmentation. Although most recent research focuses on 3D brain tumor segmentation, 2D segmentation remains important for computationally efficient pipelines, preliminary analyses, and educational purposes. Despite these advances, segmentation models are prone to errors, especially near tumor boundaries, where irregular shapes and low contrast make precise delineation challenging.
Investigating uncertainty estimation in 2D segmentation offers valuable insights into model behavior and error localization, with implications that extend to both 2D and 3D segmentation tasks.

Uncertainty estimation has been proposed as a means to identify regions where the model’s predictions may be unreliable, with the potential to guide clinicians in reviewing or refining the segmentation output. Monte Carlo (MC) Dropout is a widely used technique for estimating model uncertainty by leveraging dropout layers during inference to generate multiple stochastic forward passes. 

In this work, we investigate the relationship between the resulting variance across predictions obtained from MC Dropout, and segmentation errors in the context of 2D brain tumor segmentation. Specifically, we explore whether uncertainty estimates obtained this way can effectively highlight error-prone regions at tumor boundaries, which are critical areas for clinical decision-making. We further evaluate this relationship under different data augmentation settings to assess the robustness of uncertainty estimates and perform statistical analyses to determine the significance and practical relevance of observed correlations.

The major contributions of this work are given below:
\begin{enumerate}[label=(\roman*)]
 \item We quantify the correlation between variance-based uncertainty and segmentation error at both global and boundary levels in 2D brain tumor MRI segmentation 
\item We analyze the effect of data augmentation on uncertainty–error relationships, and
 \item We perform statistical tests to assess the robustness and significance of these correlations
\end{enumerate}

\section{Literature Survey}
In recent years, there has been increasing interest in developing trustworthy AI systems that incorporate explainability, robustness, and accountability principles to ensure that automated predictions are interpretable and reliable in clinical settings \cite{b1,b2}. In particular, uncertainty estimation has been proposed as a tool to complement segmentation predictions by allowing clinicians to identify regions where the model’s confidence is low. When uncertainty is high, clinicians can integrate additional diagnostic tests, patient history, or manual review to mitigate potential errors.
Uncertainty estimation in deep learning is broadly classified into epistemic uncertainty, which arises from limited knowledge about the model or data distribution, and aleatoric uncertainty, which stems from inherent noise in the data \cite{b3}. Epistemic uncertainty can be modelled through Bayesian frameworks or ensembles of models and is theoretically reducible with more data or improved model structure.

Uncertainty estimation in medical image analysis has been comprehensively reviewed by Zou et al. (2022) \cite{b4}, who classified approaches into Bayesian and non-Bayesian categories and discussed their applications in segmentation, detection, and classification tasks. Among these, Monte Carlo (MC) Dropout \cite{b5} has become one of the most widely adopted techniques, serving as an efficient approximation to Bayesian neural networks. By enabling dropout at inference, MC Dropout generates multiple stochastic predictions and estimates uncertainty with minimal additional computational cost \cite{b6,b7}. Numerous studies have extended MC Dropout for medical imaging tasks, including lesion detection and segmentation of polyps, as well as semi-supervised learning in cardiac and neuroimaging domains \cite{b8,b9,b10}.

However, despite its popularity, MC Dropout has been shown to suffer from several limitations. Recent studies have highlighted that uncertainty estimation methods, including MC Dropout, are poorly calibrated at finer scales and may fail to highlight error-prone regions reliably. Fuchs et al. (2022) \cite{b11} demonstrated that MC Dropout produced poorly calibrated uncertainty maps in brain tumor segmentation, especially near boundaries where errors are most likely. Similarly, Mehrtash et al. (2020) \cite{b12} found that MC Dropout’s calibration improvements were inconsistent across datasets and inferior to ensemble methods in both calibration and out-of-distribution detection. These findings suggest that dropout-based uncertainty estimation may not be sufficient for practical error localization in medical imaging.

Despite existing work highlighting the limitations of MC Dropout in uncertainty estimation for brain tumor segmentation, these studies have primarily relied on dataset-level metrics such as calibration errors or visual uncertainty maps, and have focused largely on volumetric (3D) data.  Crucially, these studies have not distinguished between uncertainty representations (variance, entropy, or mutual information) leaving it unclear whether the limitations observed are inherent to MC Dropout or specific to the choice of uncertainty metric. To the best of our knowledge, no prior work has systematically quantified the relationship between variance-based MC Dropout uncertainty and segmentation error at the per-image or boundary level in 2D segmentation tasks. Specifically, statistical analyses such as Pearson correlation and Spearman correlation between uncertainty and error, or paired t-tests comparing different augmentation settings, have not been employed to rigorously evaluate the practical utility of variance-based MC Dropout for boundary error localization. This study addresses that gap by providing a detailed, image-wise uncertainty-error analysis in 2D brain tumor segmentation across four augmentation scenarios, with a specific focus on variance as the uncertainty representation.

\section{Methodology}
\begin{figure}[htbp]
\centerline{\includegraphics[scale = 0.37]{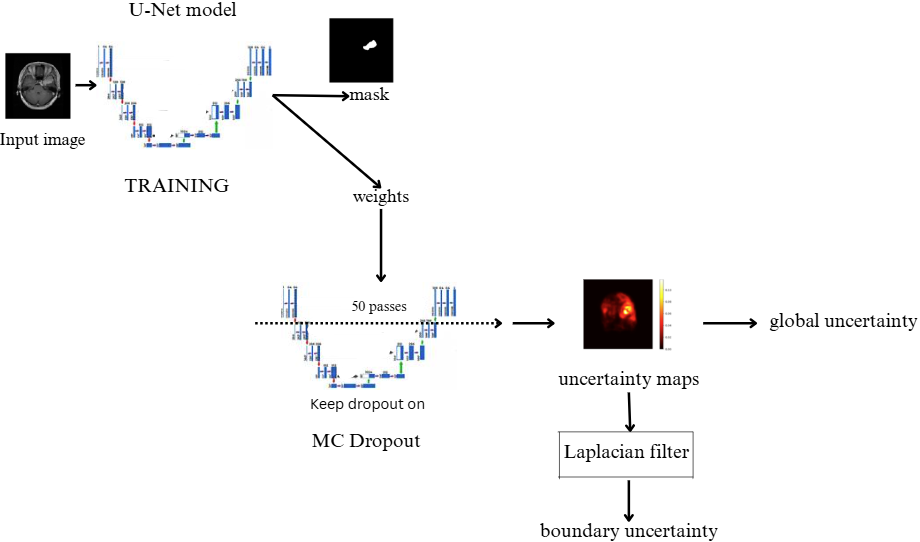}}
\caption{Workflow diagram}
\label{fig1}
\end{figure}

All experiments were conducted on a single NVIDIA T4 GPU in Google Colab with memory growth enabled to prevent full GPU memory pre-allocation. Training used tf.distribute.MirroredStrategy, which defaults to a single device in this environment and does not affect determinism. To control randomness, we fixed seeds for Python, NumPy, and TensorFlow, but did not enforce deterministic cuDNN kernels; thus, exact bit-wise reproducibility is not guaranteed. The global batch size was set to 8.

The primary objective of this study was to investigate how the uncertainty correlation with segmentation error changes across different data augmentation strategies. We trained a standard U-Net architecture using four augmentation settings—horizontal flipping, rotation, scaling, and no augmentation—evaluated independently. Although segmentation metrics such as precision, recall, and Intersection over Union (IoU) were computed to describe model performance, the main focus was on analyzing the relationship between uncertainty and error.

The trained weights from each augmentation setting were used to perform MC Dropout-based uncertainty estimation. During inference, dropout was kept active and 50 stochastic forward passes were performed per image. Pixel-wise variance across these passes was computed as the uncertainty estimate, representing one specific operationalization of epistemic uncertainty via MC Dropout. This variance-based uncertainty was then correlated with pixel-wise segmentation error to assess how well it reflects model inaccuracies under each augmentation setting, enabling a direct comparison of uncertainty-error relationships across conditions.

\subsection{Dataset}
For this study, a brain tumor dataset containing 3064 T1- weighted contrast-enhanced MRI images was used. The dataset originates from the work of Jun Cheng, who collected them from Nanfang Hospital and Tianjin Medical University in China, between 2005 and 2010 \cite{b13,b14}. The dataset comprises images from 233 patients and includes 708 cases of meningioma, 1426 cases of glioma, and 930 cases of pituitary tumors.  According to the original source, tumor boundaries were manually annotated by three experienced radiologists. While the original dataset is hosted on Figshare, we used a version available on Kaggle \cite{b15}, where the scans and corresponding binary masks were provided as 2D images of dimension 256×256 pixels. To assess class imbalance, we estimated the proportion of tumor-labelled pixels in the training masks. Over 100 randomly sampled training batches, tumor pixels comprised approximately 1.66\% of all pixels, confirming strong class imbalance.

\subsection{Preprocessing}
The image–mask pairs from the dataset were first converted to grayscale and resized to a fixed dimension of 256×256 pixels. Bilinear interpolation was used for resizing the images, while nearest-neighbor interpolation was applied to the tumor segmentation masks. The latter preserves the discrete 0/1 label values and avoids introducing gray pixels at the boundaries, which can occur with bilinear interpolation.

The images and masks were then converted to NumPy arrays and normalized by dividing by 255. The image-masks were then cast to tf.float32. Optional data augmentation was applied on-the-fly during training. Four augmentation settings were studied individually: no augmentation (baseline), horizontal flip, rotation, and scaling. The details of each augmentation settings are summarized in Table 1.

The dataset was split into training, validation, and test sets in a 60:20:20 ratio. To ensure consistency and prevent data leakage across experiments, the same splits were used for all augmentation settings. Images were first sorted and then split using a fixed random seed (seed = 42) in each notebook. As the image-mask pairs remained in the same order and the splitting process was deterministic, the resulting datasets were identical across all experiments, eliminating dataset variability. This ensures that any observed differences in model performance or uncertainty–error relationships are attributable solely to the augmentation methods, rather than variations in the dataset splits.
\begin{table}[h]
    \centering
    \caption{Details of Data Augmentation techniques applied}
    \label{tab:data_augmentation}
    \begin{tabular}{l c l}
        \toprule
        \textbf{Technique} & \textbf{\% of training dataset} & \textbf{Parameters} \\
        \midrule
        Horizontal Flip & 50\% & none \\
        Rotation & 50\% & Angle: \SI{+-15}{\degree} \\
        Random Scaling & 50\% & Range: 0.8 - 1.2 \\
        \bottomrule
    \end{tabular}
\end{table}

\subsection{Model Architecture}

We adopted the classic U-Net architecture (Ronneberger et al., 2015) \cite{b16} as the baseline model for tumor segmentation. The encoder–decoder structure with skip connections was retained without major modifications; hence, the architecture can be considered a vanilla U-Net.  Each convolutional block consisted of two convolutional layers with ReLU activation, followed by max pooling in the encoder and transposed convolutions in the decoder. To adapt the model to our task, we introduced dropout layers after each pooling and upsampling operation (to enable Monte Carlo Dropout uncertainty estimation) and employed Focal Loss (Lin et al., 2017) \cite{b17} instead of the conventional cross-entropy, to address the strong class imbalance observed in the dataset. This loss function downweights easy background pixels and emphasises hard examples near the tumor boundary. The final layer was a 1×1 convolution with sigmoid activation, producing a binary segmentation mask. The network had approximately 34.5M trainable parameters.

The U-Net model was trained using the Adam optimizer (learning rate= $1 \times 10^{-4}$) with focal loss as the loss function. Accuracy, precision, and recall were monitored as evaluation metrics during training. Because of the nature of the task at hand, accuracy metric obtained was ignored and instead emphasis was placed on precision and recall metrics. To address class imbalance and emphasize correct tumor detection, the validation recall metric was used for early stopping (patience = 7, min delta = 0.005). Training was capped at 70 epochs, and the model weights corresponding to the best validation recall were restored for final evaluation. The model with the best validation recall was restored using the early stopping callback; intermediate checkpointing was omitted to reduce training overhead.
During training, we monitored loss, precision, and recall for both training and validation sets. These metrics were plotted together as training curves and archived for each run. Hyperparameters, metrics and training curves were logged for each run and can be found in the author’s GitHub repository. After training, segmentation performance was evaluated quantitatively using the metrics described below.

\begin{table}[h!]
    \centering
    \caption{Model Hyperparameters and Training Configuration}
    \label{tab:model_config}
    \begin{tabular}{p{3cm} l l}
        \toprule
        \textbf{Category} & \textbf{Hyperparameter} & \textbf{Value} \\
        \midrule
        \multirow{8}{*}{\textbf{Model Architecture}} & Input Shape & (256, 256, 1) \\
        & Kernel Size (Initial) & (3, 3) \\
        & Kernel Size (Subsequent) & (3, 3) \\
        & Activation function & ReLU \\
        & Kernel Initializer & He Normal \\
        & Dropout Rate & 0.3 \\
        & Final Activation & Sigmoid \\
        \midrule
        \multirow{3}{*}{\textbf{Loss and Metrics}} & Loss function & Focal Loss \\
        & Optimizer & Adam \\
        & Training metrics & Accuracy \\
        \midrule
        \multirow{3}{*}{\textbf{Training Parameters}} & Batch Size & 8 \\
        & Learning Rate & $1 \times 10^{-4}$ \\
        & No. of epochs & 70 \\
        \bottomrule
    \end{tabular}
\end{table}

\subsection{Metrics for evaluating segmentation accuracy}
Segmentation quality was evaluated using both per-image and global metrics.

\subsubsection{Intersection-over-Union (IoU)}

Per-image segmentation performance was quantified using the Intersection-over-Union (IoU) metric over the test set. To avoid division by zero in cases where both ground-truth and predicted masks contained no positive pixels, a small constant ($\varepsilon = 1 \times 10^{-6}$) was added to both numerator and denominator. 

\begin{equation}
IoU = \frac{TP + \varepsilon}{TP + FP + FN + \varepsilon}
\end{equation}

Where TP, FP and FN denote the number of true positive, false positive, and false negative pixels, respectively, and $\varepsilon = 1 \times 10^{-6}$ prevents division by zero.

\subsubsection{Precision and Recall}
Pixel-level performance was further assessed using precision and recall, defined as:
 \begin{equation}
\text{Precision} = \frac{TP}{TP + FP + \varepsilon}, \quad
\text{Recall} = \frac{TP}{TP + FN + \varepsilon}
\end{equation}

where TP, FP, and FN denote the number of true positive, false positive, and false negative pixels, respectively, and $\varepsilon = 1 \times 10^{-6}$ prevents division by zero. 
During training, global precision and recall were monitored using tf.keras.metrics, while post-training evaluation employed the per-image mean IoU and its standard deviation. Uncertainty maps generated via Monte Carlo Dropout were further analyzed to examine correlations between predicted uncertainty and segmentation errors.

\subsection{Uncertainty Estimation }
To quantify model confidence in segmentation predictions, Monte Carlo (MC) Dropout [5] was employed at inference. Dropout layers, introduced after each encoder and decoder block of the U-Net, were activated during test time to perform T = 50 stochastic forward passes per image. For each pixel, the mean prediction across these passes was used as the final segmentation probability, while the pixel-wise variance across samples was used as the uncertainty estimate, representing the variance-based operationalization of epistemic uncertainty studied in this work.

Two complementary approaches were used to analyze the uncertainty maps thus obtained:

\subsubsection{Overall (total) uncertainty}
\begin{itemize}
    \item For each image, the mean of the per-pixel uncertainty map was computed, yielding a single scalar representing the overall model uncertainty for the image.
    \item This metric captures the general confidence of the model across the entire image.
\end{itemize}

\subsubsection{Boundary-specific uncertainty}
\begin{itemize}
    \item Tumor boundaries were extracted from the predicted binary mask using a Laplacian edge detection filter.
    \item Uncertainty values at the boundary pixels were then analyzed to compute:
    \begin{itemize}
            \item Mean boundary uncertainty
            \item Standard deviation of boundary uncertainty
            \item Maximum boundary uncertainty
            \item Boundary length (number of boundary pixels)
        \end{itemize}
    \item These metrics specifically capture uncertainty in regions most prone to segmentation errors, as errors are typically concentrated along tumor edges.
\end{itemize}

The resulting uncertainty metrics were stored per image and later used for correlation analysis with segmentation errors.

\subsection{Uncertainty - Error Correlation}
To investigate whether regions of high model uncertainty correspond to actual segmentation errors, we first generated per-pixel error maps by thresholding the predicted probability maps at 0.5 and comparing the resulting binary masks against the ground truth. Pixels where prediction and ground truth disagreed were marked as errors. We then quantified the relationship between model uncertainty and these error maps using both Pearson correlation coefficients (to assess linear dependence) and Spearman rank correlations (to capture monotonic but potentially non-linear relationships). For each test image, the per-pixel uncertainty and error maps were flattened, and correlations were computed independently. To specifically assess performance at tumor interfaces (regions that are particularly error-prone), we repeated the analysis by restricting pixels to those lying along the predicted tumor boundaries, identified using a Laplacian-based edge filter. This enabled a focused evaluation of whether the uncertainty estimate chosen provides meaningful cues for boundary error localization, beyond global correlations across the entire image.

Furthermore, to assess whether differences in uncertainty–error correlations across augmentation settings were statistically significant, we conducted paired statistical tests on the per-image correlation values. Specifically, we applied the paired t-test to compare mean correlations, and the Wilcoxon signed-rank test as a non-parametric alternative that does not assume normality of the differences. This dual approach ensured that our conclusions regarding statistical significance were robust to distributional assumptions.

\section{Results and Discussion}

\subsection{Segmentation Performance}
The segmentation performance across different augmentations is provided in Table 3. As seen in the table, the performance is fairly stable across different augmentations.
\begin{table}[h!]
    \centering
    \caption{Segmentation Performance Across Different Augmentation Techniques}
    \label{tab:segmentation_performance}
    
    \setlength{\tabcolsep}{2pt}

    \begin{tabular}{l *2{S[table-format=1.3]} S[table-format=1.4(4), separate-uncertainty = true]} 
        \toprule
        \textbf{Augmentation type} & {\textbf{Global Precision}} & {\textbf{Global Recall}} & {\textbf{Per image mean IoU}} \\
        \midrule
        No augmentation & 0.856 & 0.779 & 0.6910(2633) \\
        Horizontal Flip & 0.875 & 0.713 & 0.6338(2883) \\
        Rotation & 0.871 & 0.736 & 0.6556(2783) \\
        Scaling & 0.884 & 0.742 & 0.6730(2802) \\
        \bottomrule
    \end{tabular}
    
    \setlength{\tabcolsep}{6pt}
\end{table}

\subsection{Uncertainty Analysis}

The results from the uncertainty analysis across different augmentations are mentioned in Table 4. As seen in Table 4, scaling showed the least mean total uncertainty value. To further interpret model confidence, we generated per-image visualizations including predicted masks and per-pixel uncertainty maps. For clarity in the heatmap, a Gaussian filter ($\sigma = 1$) was applied to the per-pixel uncertainty maps. This smoothing was used only for visualization purposes; all quantitative analyses, including uncertainty metrics and correlation with segmentation errors, were computed on the raw, unsmoothed uncertainty values. These visualizations, as shown in Figure 2, show that the predicted uncertainty for the same input image across different augmentation settings shows visible differences. While there was visual difference between uncertainty maps, quantitative correlation with error remained low and constant across different augmentations, as shown in subsection IV (C). This indicates that per-image uncertainty maps derived from MC Dropout variance are not reliable indicators of error-prone regions in this setting.

\begin{table*}[t] 
    \centering
    \caption{Results from Uncertainty Analysis Across Different Augmentation Settings}
    \label{tab:uncertainty_results}
    
    \setlength{\tabcolsep}{6pt} 
    
    \begin{tabular}{l *4{S[table-format=0.4]}} 
        \toprule
        \textbf{Augmentation type} & {\makecell{\textbf{Average} \\ \textbf{Total uncertainty}}} & {\makecell{\textbf{Mean} \\ \textbf{boundary uncertainty}}} & {\makecell{\textbf{Max} \\ \textbf{boundary uncertainty}}} & {\makecell{\textbf{Std} \\ \textbf{boundary uncertainty}}} \\
        \midrule
        No augmentation & 0.0034 & 0.0590 & 0.1089 & 0.0192 \\
        Horizontal flip & 0.0044 & 0.0382 & 0.0656 & 0.0101 \\
        Rotation & 0.0041 & 0.0457 & 0.0763 & 0.0117 \\
        Scaling & 0.0027 & 0.0549 & 0.0985 & 0.0167 \\
        \bottomrule
    \end{tabular}
\end{table*}

\begin{figure}[htbp]
\centerline{\includegraphics[scale = 0.2]{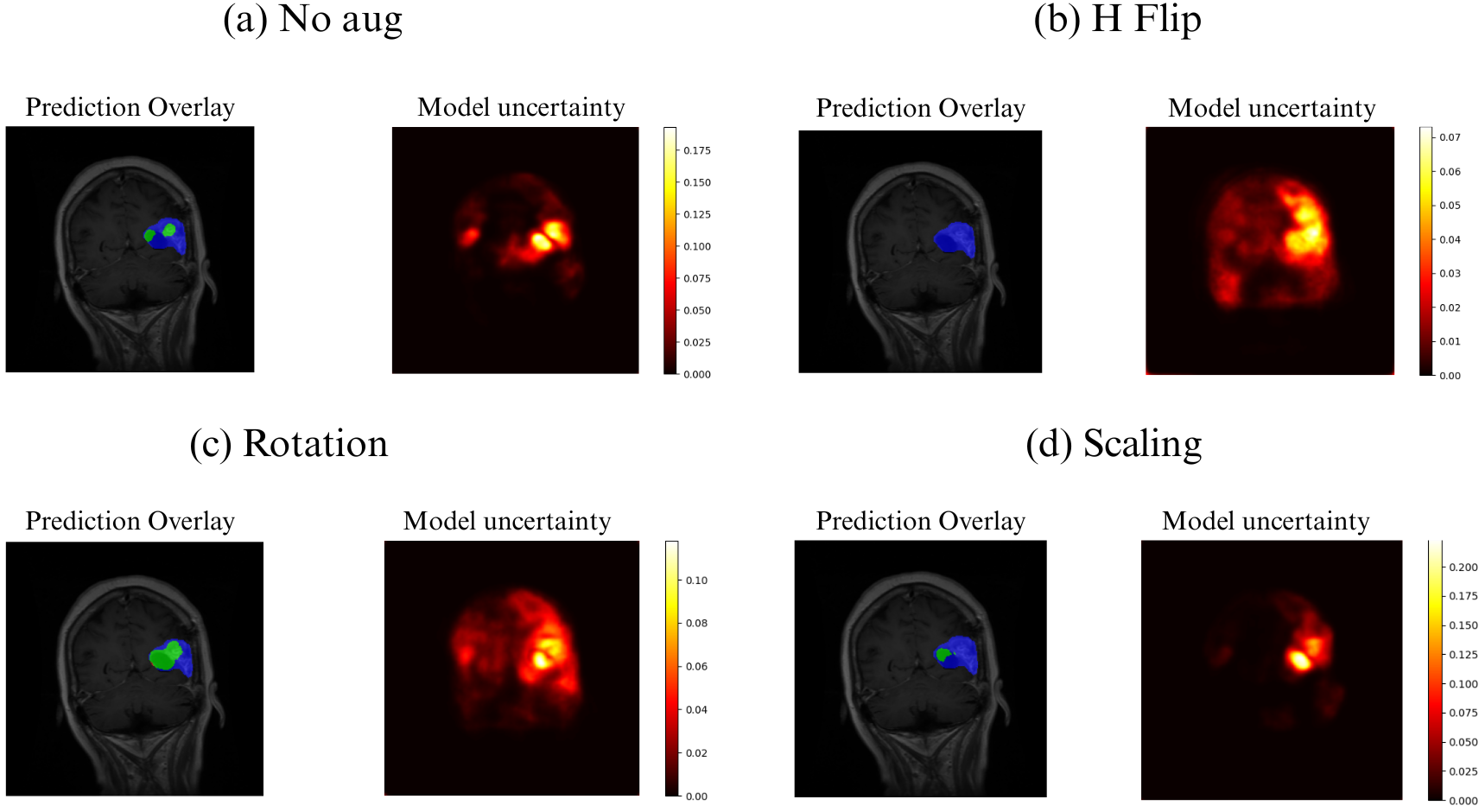}}
\caption{Predicted segmentation mask and uncertainty maps across different settings (Green = Predicted mask, Blue = Ground truth); Visual comparison shows visible differences in uncertainty maps between augmentations}
\label{fig2}
\end{figure}
 \subsection{ Uncertainty-Error Correlation}

To evaluate how well model uncertainty, obtained as variance across runs, reflected segmentation inaccuracies, we computed both Pearson’s correlation coefficient ($r$) and Spearman’s rank correlation ($\rho$) between per-image uncertainty measures and segmentation errors, under each augmentation setting.
Table 6 summarizes the correlation values.

(a)	Global correlation: Across all four settings, global correlations were modest: Pearson $r$
 ranged from ~0.30 to 0.37. Scaling produced the highest correlation (0.3781), while horizontal flip yielded the lowest (0.3014). Spearman correlation values were generally lower, ranging between 0.1115 and 0.1198, indicating weak monotonic relationships.
Although paired t-tests on the global correlation distributions yielded $p < 0.001$, the absolute differences between settings were small ($<0.07$). This indicates that while the differences are statistically detectable, they are unlikely to be of practical significance.

(b)	Boundary correlation: At the tumor boundary, Pearson correlations were close to zero, ranging between \textminus 0.0006 and 0.0458, while Spearman correlations similarly hovered around zero, between \textminus 0.0054 and 0.0453. This suggests that uncertainty obtained from variance across runs had little to no association with segmentation error at boundary regions.
\begin{table*}[htbp]
    \centering
    \caption{Statistical Tests on Mean Uncertainty–Error Correlations Between Augmentation Conditions}
    \label{tab:stat_full_comparison}
    \begin{tabular}{l *3{S[table-format=1.4]} S[table-format=2.4] S[table-format=1.4e-2] S[table-format=1.4e-2]} 
        \toprule
        \textbf{Comparison} & 
        {\makecell{\textbf{Mean Overall} \\ \textbf{Corr (A)}}} & 
        {\makecell{\textbf{Mean Overall} \\ \textbf{Corr (B)}}} & 
        {$\boldsymbol{\Delta}$ Mean} & 
        {\makecell{\textbf{p-value} \\ \textbf{(t-test)}}} & 
        {\makecell{\textbf{p-value} \\ \textbf{(Wilcoxon)}}} \\
        \midrule
        No aug v/s horizontal flip & 0.3365 & 0.3014 & +0.0351 & \num{1.4220e-10} & \num{2.1080e-12} \\
        No aug v/s rotation & 0.3365 & 0.3106 & +0.0259 & \num{1.4603e-06} & \num{8.4130e-07} \\
        No aug v/s scaling & 0.3365 & 0.3781 & -0.0416 & \num{4.6668e-14} & \num{9.3600e-21} \\
        \bottomrule
    \end{tabular}
\end{table*}
\begin{table}[h!]
    \centering
    \caption{Pearson and Spearman Rank Correlation Table}
    \label{tab:correlation_results}
    \setlength{\tabcolsep}{3pt} 
    \begin{tabular}{l *4{S[table-format=-1.4]}} 
        \toprule
        \textbf{Augmentation type} & \multicolumn{2}{c}{\textbf{Pearson $r$}} & \multicolumn{2}{c}{\textbf{Spearman $\rho$}} \\
        \cmidrule(lr){2-3} \cmidrule(lr){4-5} 
        & {\makecell{Global}} & {\makecell{Boundary}} & {\makecell{Global}} & {\makecell{Boundary}} \\
        \midrule
        No augmentation & 0.3365 & -0.0006 & 0.1115 & -0.0054 \\
        Horizontal flip & 0.3014 & 0.0369 & 0.1198 & 0.0349 \\
        Rotation & 0.3106 & 0.0458 & 0.1171 & 0.0453 \\
        Scaling & 0.3781 & 0.0195 & 0.1136 & 0.0171 \\
        \bottomrule
    \end{tabular}
    \setlength{\tabcolsep}{6pt}
\end{table}

As shown in Figure 2, regions of high uncertainty do not consistently coincide with actual segmentation errors; high variance is observed even in regions where the model predicted correctly. This qualitatively supports the quantitative finding that variance-based uncertainty is not a reliable indicator of error-prone regions in this setting.

Table 5 presents the pairwise comparison of mean uncertainty–error correlations between the baseline (no augmentation) and each augmentation setting. Although the absolute differences in mean correlation ($\Delta \text{Mean}$) were small ($<0.05$), both the paired t-test and Wilcoxon signed-rank test yielded highly significant p-values ($p<0.001$) for all comparisons. This indicates that while augmentation-induced differences are statistically detectable due to the size and consistency of the test set, their magnitude is negligible, suggesting limited practical significance.
Overall, the results indicate that MC Dropout variance-based uncertainty exhibits only weak global correlation with segmentation error, and provides little to no information for boundary error prediction. Moreover, the minimal variation across augmentation settings highlights that data augmentation has limited influence on this uncertainty–error relationship in the 2D brain tumor segmentation task studied here.

\subsection{Discussion}
The analysis indicates that variance-based MC Dropout uncertainty correlates weakly with segmentation error in the context of 2D brain tumor segmentation, with correlations being especially negligible at tumor boundaries. While global uncertainty metrics showed weak correlation (Pearson $r$ ranging from 0.3014 to 0.3781), boundary correlations were negligible ($|r| < 0.05$), suggesting that MC Dropout variance-based uncertainty estimates do not reliably highlight regions with segmentation errors in this setup.

Although paired t-tests on the global correlation distributions yielded $p < 0.001$, the absolute differences between the augmentation settings were small ($< 0.07$), implying that the differences are statistically detectable but unlikely to be of practical significance. This further reinforces the notion that augmentation variations do not meaningfully affect the relationship between uncertainty and error in this setup.

It is important to note that these findings are specific to variance-based uncertainty estimation in 2D brain tumor segmentation using a U-Net with MC Dropout, and should not be generalized to other architectures, data modalities, or uncertainty representations. Within this experimental setup, variance is a poor proxy for segmentation error, particularly at boundaries, suggesting that the choice of uncertainty representation may matter more than the underlying method itself. Predictive entropy and mutual information are promising alternatives that may better capture boundary error localization.

\section{Conclusion and Future Scope}
In this work, we evaluated the suitability of variance-based MC Dropout uncertainty for identifying segmentation errors in 2D brain tumor segmentation, and the effect of augmentation on uncertainty-error correlations. Results show weak global correlations and negligible boundary-level correlations across all augmentation settings, suggesting limited utility of variance-based uncertainty for both global and boundary error localization. Differences across augmentation conditions were statistically significant but practically negligible, indicating that data augmentation does not meaningfully influence this uncertainty–error relationship.

These findings suggest that variance is a poor proxy for segmentation error in this setting, and that the choice of uncertainty representation may matter more than the method itself. Future work should explore alternative representations such as predictive entropy and mutual information, which may better capture boundary error localization, as well as extend this analysis to other architectures and datasets to assess broader applicability.

\vspace{\baselineskip}

\textit{\textbf{Data and code availability:}}
All code and experimental configurations are publicly available at https://github.com/Saumya4321/mcd-error-correlation.

\vspace{12pt}


\begin{thebibliography}{00}
\bibitem{b1} Li, B., Qi, P., Liu, B., Di, S., Liu, J., Pei, J., Yi, J. and Zhou, B., 2023. Trustworthy AI: From principles to practices. ACM Computing Surveys, 55(9), pp.1-46. 
\bibitem{b2} Liang, W., Tadesse, G.A., Ho, D., Fei-Fei, L., Zaharia, M., Zhang, C. and Zou, J., 2022. Advances, challenges and opportunities in creating data for trustworthy AI. Nature Machine Intelligence, 4(8), pp.669-677.
\bibitem{b3} Der Kiureghian, A. and Ditlevsen, O., 2009. Aleatory or epistemic? Does it matter?. Structural safety, 31(2), pp.105-112.
\bibitem{b4} Zou, K., Chen, Z., Yuan, X., Shen, X., Wang, M. and Fu, H., 2023. A review of uncertainty estimation and its application in medical imaging. Meta-Radiology, 1(1), p.100003.
\bibitem{b5} Gal, Y. and Ghahramani, Z., 2016, June. Dropout as a Bayesian approximation: Representing model uncertainty in deep learning. In International Conference on machine learning (pp. 1050-1059). PMLR.
\bibitem{b6} Srivastava, N., Hinton, G., Krizhevsky, A., Sutskever, I. and Salakhutdinov, R., 2014. Dropout: a simple way to prevent neural networks from overfitting. The Journal of machine learning research, 15(1), pp.1929-1958.
\bibitem{b7} Lakshminarayanan, B., Pritzel, A. and Blundell, C., 2017. Simple and scalable predictive uncertainty estimation using deep ensembles. Advances in neural information processing systems, 30.
\bibitem{b8} Nair, T., Precup, D., Arnold, D.L. and Arbel, T., 2020. Exploring uncertainty measures in deep networks for multiple sclerosis lesion detection and segmentation. Medical image analysis, 59, p.101557.
\bibitem{b9} Wickstrøm, K., Kampffmeyer, M. and Jenssen, R., 2020. Uncertainty and interpretability in convolutional neural networks for semantic segmentation of colorectal polyps. Medical image analysis, 60, p.101619.
\bibitem{b10} Yu, L., Wang, S., Li, X., Fu, C.W. and Heng, P.A., 2019, October. Uncertainty-aware self-ensembling model for semi-supervised 3D left atrium segmentation. In International Conference on Medical Image Computing and Computer-Assisted Intervention (pp. 605-613). Cham: Springer International Publishing.
\bibitem{b11}	Fuchs, M., Gonzalez, C. and Mukhopadhyay, A., 2021. Practical uncertainty quantification for brain tumor segmentation. In Medical Imaging with Deep Learning.
\bibitem{b12}	Mehrtash, A., Wells, W.M., Tempany, C.M., Abolmaesumi, P. and Kapur, T., 2020. Confidence calibration and predictive uncertainty estimation for deep medical image segmentation. IEEE transactions on medical imaging, 39(12), pp.3868-3878.
\bibitem{b13} Cheng, J., Huang, W., Cao, S., Yang, R., Yang, W., Yun, Z., Wang, Z. and Feng, Q., 2015. Enhanced performance of brain tumor classification via tumor region augmentation and partition. PloS one, 10(10), p.e0140381
\bibitem{b14}	Cheng, J., Yang, W., Huang, M., Huang, W., Jiang, J., Zhou, Y., Yang, R., Zhao, J., Feng, Y., Feng, Q. and Chen, W., 2016. Retrieval of brain tumors by adaptive spatial pooling and Fisher vector representation. PloS one, 11(6), p.e0157112.
\bibitem{b15} Kaggle, Brain Tumor Segmentation dataset [Online]. Available: https://www.kaggle.com/datasets/nikhilroxtomar/brain-tumor-segmentation
[Accessed Sept. 18, 2025]
\bibitem{b16} Ronneberger, Olaf, Philipp Fischer, and Thomas Brox. U-net: Convolutional networks for biomedical image segmentation. In International Conference on Medical image computing and computer-assisted intervention, pp. 234-241. Cham: Springer international publishing, 2015
\bibitem{b17}Lin, Tsung-Yi, Priya Goyal, Ross Girshick, Kaiming He, and Piotr Dollár. Focal loss for dense object detection. In Proceedings of the IEEE international conference on computer vision, pp. 2980-2988. 2017


\end{thebibliography}
\end{document}